\documentclass{IOS-Book-Article}

\usepackage{mathptmx}
\usepackage{soul}\setuldepth{article}
\usepackage{microtype}      
\usepackage[numbers]{natbib}
\usepackage{amsmath}
\usepackage{amssymb}
\usepackage{mathtools}
\usepackage{amsthm}
\usepackage{array}
\usepackage{booktabs}
\usepackage{algorithmic}
\usepackage{multirow} 
\usepackage{threeparttable}
\usepackage[table]{xcolor}
\usepackage{graphicx} 
\usepackage{float}    
\usepackage{caption} 
\usepackage{subcaption}
\usepackage{adjustbox} 
\usepackage{wrapfig}
\usepackage{hyperref}

\def\hb{\hbox to 11.5 cm{}}
\newcommand{\header}[1]{\noindent \textit{\textbf{#1}}}

\begin{document}

\pagestyle{headings}
\def\thepage{}
\begin{frontmatter}              

\title{Mining Social Determinants of Health for Heart Failure Patient 30-Day Readmission via Large Language Model}

\markboth{}{January 2025\hb}

\author[A]{\fnms{Mingchen} \snm{Shao}},
\author[B]{\fnms{Youjeong} \snm{Kang}}
\author[B]{\fnms{Xiao} \snm{Hu}}
\author[B]{\fnms{Hyunjung Gloria} \snm{Kwak}}
\author[B]{\fnms{Carl} \snm{Yang}}
and
\author[B]{\fnms{Jiaying} \snm{Lu}\thanks{Corresponding Author: Jiaying Lu, jiaying.lu@emory.edu.}}

\runningauthor{M. Shao et al.}
\address[A]{Department of Computer Science, Emory University}
\address[B]{Nell Hodgson Woodruff School of Nursing, Emory University}

\begin{abstract}
Heart Failure (HF) affects millions of Americans and leads to high readmission rates, posing significant healthcare challenges.
While Social Determinants of Health (SDOH) such as socioeconomic status and housing stability play critical roles in health outcomes, they are often underrepresented in structured EHRs and hidden in unstructured clinical notes. This study leverages advanced large language models (LLMs) to extract SDOHs from clinical text and uses logistic regression to analyze their association with HF readmissions. By identifying key SDOHs (\textit{e.g.} tobacco usage, limited transportation) linked to readmission risk, this work also offers actionable insights for reducing readmissions and improving patient care.
\end{abstract}

\end{frontmatter}
\markboth{January 2025\hb}{January 2025\hb}

\section{Introduction}

Heart Failure (HF), the condition in which the heart fails to pump blood and oxygen to the body's organs, is a significant public health concern. Approximately 6.7 million Americans over 20 years of age have HF, and the prevalence is expected to rise to 11.4 million by 2050~\cite{Bozkurt2024}. Nearly 25\% HF patients are readmitted within 30 days of discharge, causing high healthcare cost~\cite{Muhammad2021}. While risk factors for readmission have been predominantly studied in the context of medical factors such as comorbidities~\cite{Wideqvist2021}, the role of Social Determinants of Health (SDOH) (\textit{e.g.} socioeconomic status and social support) has not been well studied~\cite{ahsan2021mimicsbdh}. 
Understanding how SDOH affect readmission among HF patients using available patient health data could provide critical insights to develop appropriate interventions, ultimately improving patient outcomes and reducing healthcare costs.

A great challenge for SDOH based readmission analysis lies in SDOH being often underrepresented or insufficiently captured in structured electronic health records (EHRs) or insurance claims~\cite{guevara2024large}. Fortunately, unstructured clinical notes (e.g. admission notes, nursing progress notes, discharge notes) contains rich, detailed narratives that convey information regarding SDOH (e.g. “pat live alone w/ pets”)~\cite{guevara2024large,chen2020social}. Extracting and analyzing this information is challenging due to the variability and free-text nature of clinical notes.
To address this, we propose to leverage the advanced large language model (LLM), which is capable of processing and understanding domain-specific text~\cite{guevara2024large}, to systemically mine SDOH. We further aim to understand the association between SDOH and HF readmission by performing odds ratio analysis using logistic regression~\cite{Sperandei2014}. This approach allows us to quantify the relative impact of factors such as socioeconomic status, housing stability, and access to transportation on readmission likelihood. By interpreting the odds ratios, we identify which SDOHs were most strongly associated with increased or decreased risk of readmission, providing actionable insights for targeted interventions.
\section{Methods}
\header{Dataset and Cohort Selection.}
For our data source, we use Medical Information Mart for Intensive Care (MIMIC-III)~\cite{johnson2016mimiciii}, a large open source healthcare database that contains information for patient hospitalized in critical care unit. 
We first identify HF patients through their recorded ICD-9 code: ``\textit{398.91, 402.01, 402.11, 402.91, 404.01, 404.03, 404.11, 404.13, 404.91, 404.93}'', and any code that starts with ``\textit{428}''. We keep two consecutive ICU admissions of HF patients as one data sample, and whether the admission date differences between the current and previous admission is less than 30 days is used as the prediction target. 
Table \ref{tab:cohort_stat} summarizes the statistics of the built cohort.
Multiple discharge notes can correspond to a single hospital admission due to caregivers writing multiple notes and adding addendum during the admission. Therefore, we remove duplicated texts and concatenate these notes together for SDOH mining. For overly lengthy combined notes, we keep mainly \textit{Social History}, \textit{Physical Exam}, and \textit{Medical History} sections that mostly likely mention SDOHs. 

\begin{table}[htbp!]
    \centering
    \caption{Statistics of our built datasets. ReAdm denotes 30-day readmission. LoS denotes Length of Stay (days).}
    \label{tab:cohort_stat}
    \resizebox{\linewidth}{!}{
    \begin{tabular}{cccccc}
    \toprule
    \#HF Pat & \#Notes & ReAdm (\%)& Median LoS (Q1-Q3) & Median Age (Q1-Q3) & Gender, Female (\%)\\
    \midrule
    2065 & 3604 & 33.19 & 7.0 (4.0--13.0) & 72.9 (62.6--81.3) & 1107 (46.39\%)\\ 
    \bottomrule
    \end{tabular}
    }
\end{table}

\header{Social Determinants of Health Mining.} 
One primary goal of our study is to mine SDOHs from textual notes. Given a patient's (discharge) note $\mathbf{T}=\{t_1, \dots, t_{|\mathbf{T}|}\}$ consist of a sequences of tokens, we leverage the LLM $f_\theta$ to extract certain SDOH $\textbf{D}=\{d_1, \dots, d_{|\mathbf{D}|}\}$ that can be either a text span from original note $\mathbf{T}$ or a paraphrase of the original text span. 
Therefore, the mining process is defined as $\mathbf{D}=f_\theta(\mathbf{T},\mathbf{P_{D}})$, where $\mathbf{P_{D}}$ denotes the SDOH specific prompt. 
To prompt the LLM, we use zero-shot prompting, a prompting method that instructs a language model to perform a task without providing specific examples or training data for that particular task~\cite{brown2020language}. The key advantage is that the learnable parameters $\theta$ of $f_{\theta}$ would not need to be updated. Instead, we utilize the checkpoint of an open source LLM, Llama-3.1-8B~\cite{grattafiori2024llama3herdmodels}, where the 8 billion parameters $\theta$ have been pre-trained with massive amount of textual corpus including biomedical literature and public clinical databases.
Specifically, we use a template of prompt to mine various SDOHs: 

\vspace{0.15cm}
\hrule
(\textit{Prompt for SDOH $\mathbf{D}_i$}) ``Can you extract the patient’s $[name(\mathbf{D}_i)]$ from the given discharge note? Please choose from $candidate([\mathbf{D}_i)]$, or `unspecified' if not mentioned in the note. \#\#Note\#\#: [$\mathbf{D}$]''.
\hrule
\vspace{0.15cm}

\noindent In the prompt, text in ``[]'' are placeholders to be filled with specific values. For instance, the filled prompt for Alcohol usage is ``Can you extract the patient’s use of alcohol from the given discharge note? Please choose from `present', `past', `never', or ‘unspecified’...''
Overall, we covered two categories of SDOHs: (1) charted in EHR: \textbf{Gender, Age, Ethnicity, Language, Marital Status, Insurance}; (2) non-charted: \textbf{Alcohol, Tobacco, Drug, Transportation, Housing, Parental, Employment, Social Support}.

\vspace{0.3cm}
\header{SDOH and HF 30-Day Readmission Association Analysis.}
We conduct a logistic regression analysis to evaluate the association between various SDOHs and 30-day readmission risk among HF patients. Let $\mathbf{X}^\intercal = [x_1,\dots, x_{|\mathbf{D}|}]$ represent the vector of predictors derived from $\mathbf{D}$ for a given patient, and $y\in \{0, 1\}$ be the binary outcome variable where $y=1$ indicates that the patient is readmitted within 30 days and $y=0$ otherwise. The logistic regression model is defined as: $P(y = 1 \mid \mathbf{X}) = \frac{1}{1 + e^{-(\beta_0 + \sum_{j=1}^{|D|} \beta_j x_j)}}$,
where $\beta_0$ is the intercept term and $\beta_j$ is the coefficient for the $j$-th predictor.
Odds ratios (ORs) are calculated for each SDOH to quantify the magnitude and direction of their effect on the likelihood of readmission. The odds ratio for a specific value $j$ of SDOH $i$ is defined as:
$OR_{i,j} = \frac{P(y=1 \mid D_{i,j})/P(y=0 \mid D_{i,j})}{P(y=1\mid D_{i,ref}/P(y=0 \mid D_{i,ref})}$,
where $D_{i,j}$ denotes a specific value $j$ of the  $i$-th SDOH (e.g., $D_{\text{Race,Black}}$).
$D_{i,\text{ref}}$ denotes the reference value for the  $i$-th SDOH (e.g., White for Race).
For each SDOH $i$, we choose a reference group $D_{i,\text{ref}}$  to serve as the baseline for comparison, so that all odds ratios represent the likelihood of readmission relative to the reference group. The subscript $j$ represents all possible values of $D_i$, excluding the reference group. Variables with an $OR > 1$ are considered to increase the risk of readmission, whereas variables with an $OR < 1$ are considered protective factors. Confidence intervals (95\%) and p-values are used to assess the statistical significance of these associations.
\section{Results}

\begin{table}[htbp!]
\centering
\vspace{-0.5cm}
\caption{LLM SDOH mining performance. Acc denotes the accuracy over all samples. Spec-Acc and \% spec denotes the accuracy and percentage of attributes NOT extracted as \textbf{unspecified}, separately.}
\vspace{-0.2cm}
\begin{subtable}[t]{0.49\linewidth}
\centering
\caption{Charted SDOHs.}
\label{tab:mine_total}
\begin{tabular}{ccc}
\toprule
 Attribute & Acc& Spec-Acc (\% Spec)\\
\midrule
 Gender & 98.27\%&99.80\% (96.75\%)\\
\hline
 Age (MAE$\downarrow$)& \multicolumn{2}{c}{3.08 ( 97.95\%)}\\
\hline
 Ethnicity & 8.3\%& 73.91\% (6.38\%)\\
\hline
 Language & 2.38\%&50.13\% (4.94\%)\\
\hline
Marital Status & 42.08\%& 70.21\% (58.99\%)\\
\hline
Insurance &2.63\%&45.97\% (5.85\%)\\
\bottomrule
\end{tabular}
\end{subtable}
\begin{subtable}[t]{0.46\linewidth}
\caption{Non-charted SDOHs.}
\label{tab:mine_uncharted}
\begin{tabular}{ccc}
\toprule
Attribute & Acc& Spec-Acc (\% Spec)\\
\midrule
Alcohol & 69.44\%& 73.02\% (11.21\%)\\
\hline
Tobacco & 79.19\%& 81.15\% (12.51\%)\\
\hline
Drug & 67.46\% & 57.39\% (12.40\%)\\
\hline
Transportation & 95.71\%& 0.00\% (0.08\%)\\
\hline
Housing & 87.14\%& 0.00\% (0.02\%)\\
\hline
Parental & 68.57\%& 0.00\% (0.58\%) \\
\hline
Employment & 68.57\%& 7.14\% (0.39\%)\\
\hline
Social Support & 44.29\%& 11.32\% (1.47\%)\\
\bottomrule
\end{tabular}
\end{subtable}
\end{table}
\header{SDOH Mining Experiments.}
After obtaining the mined results, we verify the charted SDOH attributes against the corresponding EHR data (Table~\ref{tab:mine_total}) and the uncharted SDOH attributes against human annotated dataset~\cite{ahsan2021mimicsbdh,guevara2024large} (Table~\ref{tab:mine_uncharted} ). For categorical attributes, we use accuracy as the evaluation metric, while for the continuous variable ``Age'', we use the mean absolute error (MAE). 
Given the under-documented characteristic of SDOH in clinical notes~\cite{guevara2024large}, we further report two additional metrics, Spec-Acc (accuracy of specified SDOHs) and \% Spec (percentage of notes specifying SDOHs over all notes) to focus on the notes that indeed mentioned SDOH attributes (excluding those extracted as ``unspecified'' samples). 
For attribute `Age,' which is evaluated using MAE, no distinction is necessary since MAE inherently ignores null values, which are those marked as ``unspecified." 

\begin{table}[htbp!]
\centering
\vspace{-0.3cm}
\caption{Significant Predictors of HF Readmission}
\label{tab:significant_predictors}
\begin{tabular}{lcc}
\toprule
Variable & OR (95\% CI) & p-value \\ 
\midrule
Age & 1.01 (1.00--1.01) & $< 0.01$ \\
Ethnicity (Unspecified) & 1.73 (1.33--2.64) & $< 0.01$ \\
Insurance (Medicare) & 5.23 (0.93--29.29) & $< 0.05$ \\
Tobacco (Past) & 1.20 (0.99--1.46) & $< 0.05$ \\
Transportation (Limited) & 2.07 (0.88--4.87) & $< 0.05$ \\
Social Support (Positive) & 0.58 (0.33--1.00) & $< 0.05$ \\
\bottomrule
\end{tabular}
\end{table}

\header{Logistic Regression Statistical Inference.}
The results of the logistic regression, highlighting 
significant predictors of HF readmission, are shown in Table \ref{tab:significant_predictors}. Age is identified as a significant predictor, indicating a slight increase in the likelihood of readmission with increasing age. Ethnicity, specifically when documented as ``unspecified'', is associated with a higher risk of readmission. Among insurance types, Medicare is significantly associated with a markedly increased risk of readmission. Past tobacco use and transportation resource limitation are also significant predictors. Finally, social support is identified as a significant factor associated with a reduced likelihood of readmission.

\section{Discussion}

The high Spec-Acc for extracted result, especially for patient gender, marital status, alcohol, and tobacco use, as well as the low MAE for age demonstrates the LLM's effectiveness in extracting structured information from unstructured text. The significantly low Spec-Acc for other uncharted SDOH is likely due to the human-annotated dataset containing very few instances available for verification. Additionally, these instances represent a small proportion of the entire cohort. Consequently, the low accuracy may not accurately reflect the true performance of the LLM. 

Logistic regression analysis further highlights key factors influencing 30-day readmission. Older individuals face a higher risk of readmission, aligning with prior studies~\cite{CHEN1999605} that underscore the increased heart failure risk in this population. Unspecified ethnicity warrants further investigation, potentially reflecting data limitations, reporting biases, or reluctance due to historical bias. Medicare insurance is strongly associated with readmission, suggesting unique challenges faced by this patient population. Behavioral factors like past tobacco use, and social determinants, including transportation barriers and lack of social support, significantly contribute to readmission risk, which align with previous studies~\cite{syed2013transportation,chen2020social}. Addressing these issues through targeted interventions could help mitigate readmission risks.

One limitation of our analysis is that we focus solely on discharge notes to extract SDOH information, excluding other potentially valuable sources such as nursing and physician notes. These additional clinical notes might provide complementary insights into patient circumstances and could enhance the comprehensiveness of future analyses. Addressing this limitation in subsequent research could improve the robustness of SDOH identification and its predictive power for HF patient outcomes.
\section{Conclusion}
Our training-free LLM-based approach enables SDOH mining from unstructured clinical notes. 
The effectiveness of our mining method is justified by the great performance over annotated data. Furthermore, we conduct downstream 30-day readmission correlation analysis to partially support the quality of mined SDOH. Multiple identified significant predictors of HF readmission, including old age, past tobacco usage, transportation limitation, and lack of social support,  align with existing clinical findings.
In the future, we plan to (1) extend our LLM-based mining approach to cover more aspects of SDOHs from various types of clinical notes, and (2) explore advanced causal relationship mining to identify novel SDOH to provide actionable insights for reducing HF readmission rate. 
Observing the limited numbers of notes containing certain SDOHs, we advocate for clinicians to document these factors more consistently to enhance their utility for downstream analysis and intervention planning. 



\small
\bibliographystyle{unsrt}
\bibliography{mybib}

\end{document}